\documentclass{article}

\usepackage[final]{neurips_2021}

\usepackage[dvipsnames]{xcolor}

\usepackage[utf8]{inputenc} 
\usepackage[T1]{fontenc}    
\usepackage{hyperref}       
\usepackage{url}            
\usepackage{booktabs}       
\usepackage{amsfonts}       
\usepackage{nicefrac}       
\usepackage{microtype}      
\usepackage{xcolor}         
\usepackage{listings}
\usepackage{subcaption}
\usepackage{caption}

\usepackage{tcolorbox}
\usepackage[scaled=.95]{inconsolata}
\definecolor{myteal}{RGB}{110, 180, 160}

\usepackage{pythonhighlight}

\title{\texttt{FasterAI}: A Lightweight Library for Creating Sparse Neural Networks}

\author{%
  Nathan Hubens \\
  ISIA Lab, University of Mons, Belgium\\
  Artemis, IP Paris, France\\
  \texttt{nathan.hubens@umons.ac.be} \\
}

\begin{document}

\maketitle

\begin{abstract}
\texttt{FasterAI} is a PyTorch-based library, aiming to facilitate the utilization of deep neural networks compression techniques such as sparsification, pruning, knowledge distillation, or regularization. The library is built with the purpose of enabling quick implementation and experimentation. More particularly, compression techniques are leveraging \texttt{Callback} systems of libraries such as fastai  and Pytorch Lightning to bring a user-friendly and high-level API. The main asset of \texttt{FasterAI} is its lightweight, yet powerful, simplicity of use. Indeed, because it was developed in a very granular way, users can create thousands of unique experiments by using different combinations of parameters. In this paper, we focus on the sparsifying capabilities of \texttt{FasterAI}, which represents the core of the library. Performing sparsification of a neural network in \texttt{FasterAI} only requires a single additional line of code in the traditional training loop, yet allows to perform state-of-the-art techniques such as Lottery Ticket Hypothesis experiments.

\end{abstract}

\section{Introduction}

\texttt{FasterAI} \cite{fasterai} is an open-source library, accessible on GitHub\footnote{\url{https://github.com/nathanhubens/fasterai}}, released under an Apache-2.0 license, and that can be installed through the \texttt{pip} package manager. It also includes extensive documentation and several tutorials \footnote{\url{https://nathanhubens.github.io/fasterai/}}.

Although \texttt{FasterAI} contains diverse neural network compression techniques such as sparsification, pruning, knowledge distillation, and regularization, we focus this paper on developing the sparsification capabilities of the library. \texttt{FasterAI}'s main goal is to provide to PyTorch \cite{pytorch} users a seamless access to sparsification techniques. Indeed, the only code required to begin sparsifying Pytorch models trained with fastai \cite{fastai} or Pytorch Lightning \cite{lightning} is the following: 

 \begin{python}
from fasterai.sparse import SparsifyCallback

sp_cb=SparsifyCallback(sparsity, granularity, context, criteria, schedule)
learner.fit(n_epochs, cbs=sp_cb)
 \end{python}

We show in this paper that, passing different combinations of arguments in the \texttt{SparsifyCallback} allows to recreate the most common pruning configurations, but also to easily customize the callback to create novel pruning techniques.

\section{The \texttt{SparsifyCallback}}

The whole power of sparsification capabilities of FasterAI lies in its \texttt{SparsifyCallback}, designed around 4 independent building blocks: \texttt{granularity}, \texttt{context}, \texttt{criteria}, and \texttt{schedule}.

In \texttt{FasterAI} documentation, we make the distinction between the process of sparsification, \textit{i.e.} making neural network's weight sparse, and pruning, \textit{i.e.} physically removing those sparse weights. To better conform to the literature, those terms may be used interchangeably in this paper, but will always designate sparse neural network models.

\texttt{FasterAI} possesses two important ways to create a sparse network: (1) the static way, by using the \texttt{Sparsifier} class, able to sparsify either a specified layer, or the whole model, (2) the dynamic way, by using the \texttt{SparsifyCallback}, that must be used in conjunction with training, and removing weights while the network is learning. Examples of usage for both methods are expressed below:

 \begin{python}
# (1) Static
sp=Sparsifier(model, granularity, context, criteria)
sp.prune_model(sparsity)
 
# (2) Dynamic
sp_cb=SparsifyCallback(sparsity, granularity, context, criteria, schedule)
learner.fit(n_epochs, cbs=sp_cb)
 \end{python}

It is important to mention that \texttt{FasterAI} does not physically remove the weights of a network, but rather creates a binary mask of the same structure as the network's weights, and applies it to either prune a weight (when the mask value is 0) or keep it unchanged (when the mask value is 1).

The \texttt{SparsifyCallback} is built around the 4 main questions that should be answered in order to completely describe a sparsification technique, with each of its arguments aiming to provide an answer to each of those. The arguments and corresponding questions are:

\begin{itemize}
    \item \texttt{granularity}: how to sparsify?
    \item \texttt{context}: where to sparsify?
    \item \texttt{criteria}: what to sparsify?
    \item \texttt{schedule}: when to sparsify?
\end{itemize}

The purpose is to decompose the sparsifying problem into 4 subproblems, related to different fields of research. By doing so, each argument can be modified independently from the others, which allows to: (1) create a vast number of opportunities and combinations for experiments and, (2) provide a unique and versatile callback, reducing the problem of implementing a novel sparsification technique to the modification of a single argument.

\subsection{\texttt{granularity}: how to sparsify?}

\begin{figure}[!htbp]
     \centering
     \begin{subfigure}[t]{0.24\textwidth}
         \centering
         \includegraphics[width=\textwidth]{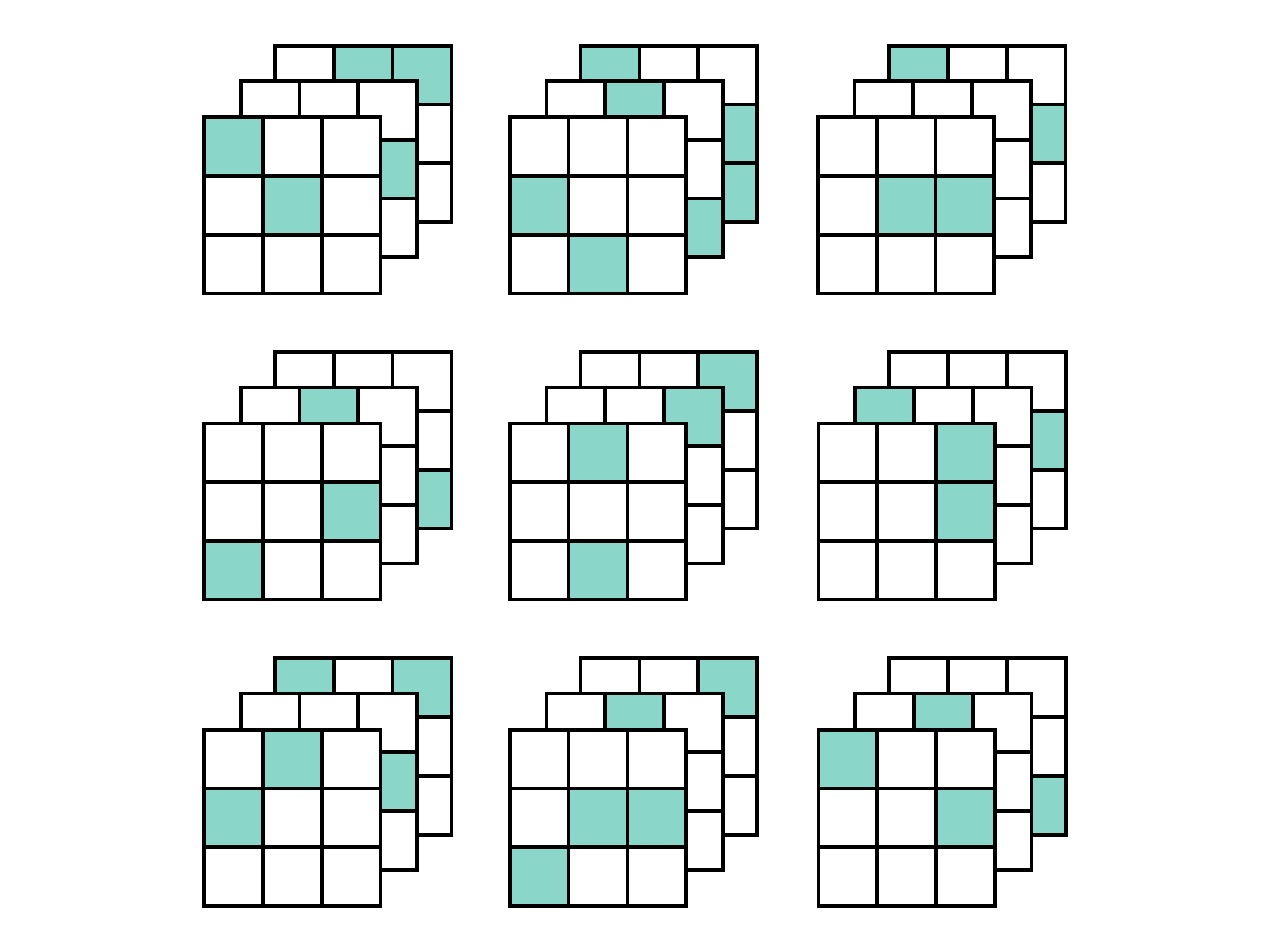}
         \caption{\texttt{\textquotesingle weight\textquotesingle}}
         \label{fig:y equals x}
     \end{subfigure}
     \hfill
     \begin{subfigure}[t]{0.24\textwidth}
         \centering
         \includegraphics[width=\textwidth]{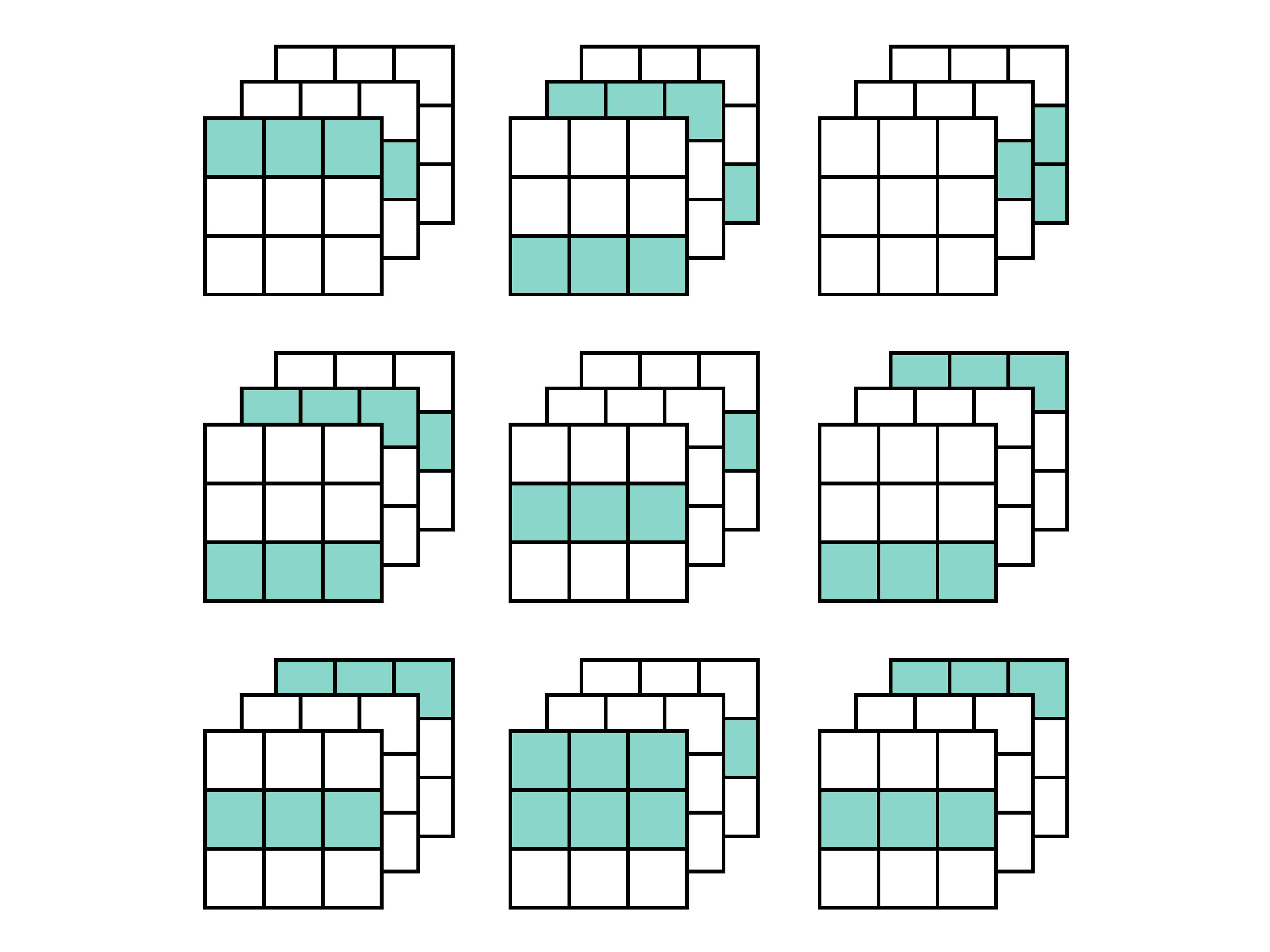}
         \caption{\texttt{\textquotesingle row\textquotesingle}}
         \label{fig:three sin x}
     \end{subfigure}
     \hfill
     \begin{subfigure}[t]{0.24\textwidth}
         \centering
         \includegraphics[width=\textwidth]{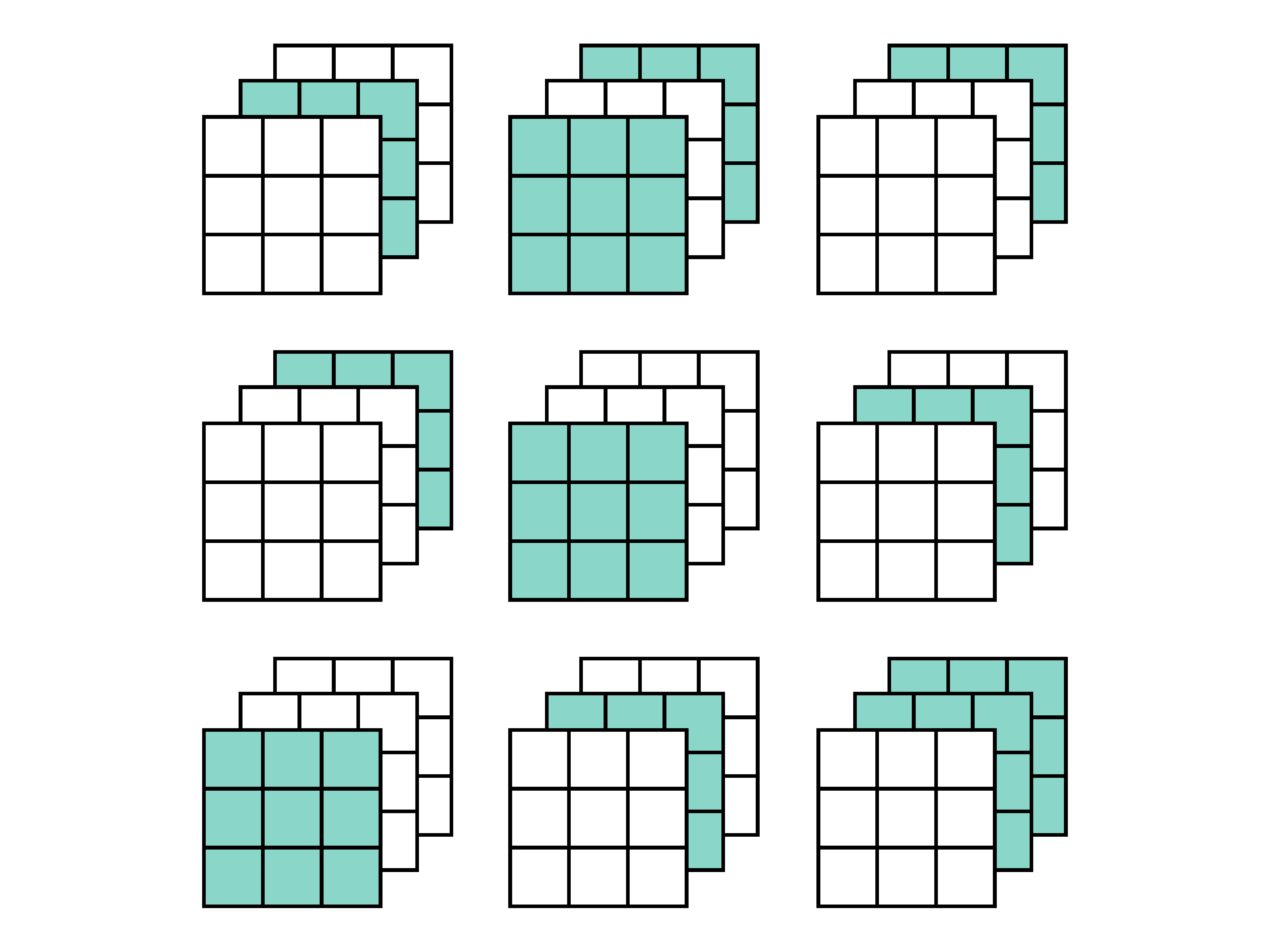}
         \caption{\texttt{\textquotesingle kernel\textquotesingle}}
         \label{fig:five over x}
     \end{subfigure}
    \hfill
     \begin{subfigure}[t]{0.24\textwidth}
         \centering
         \includegraphics[width=\textwidth]{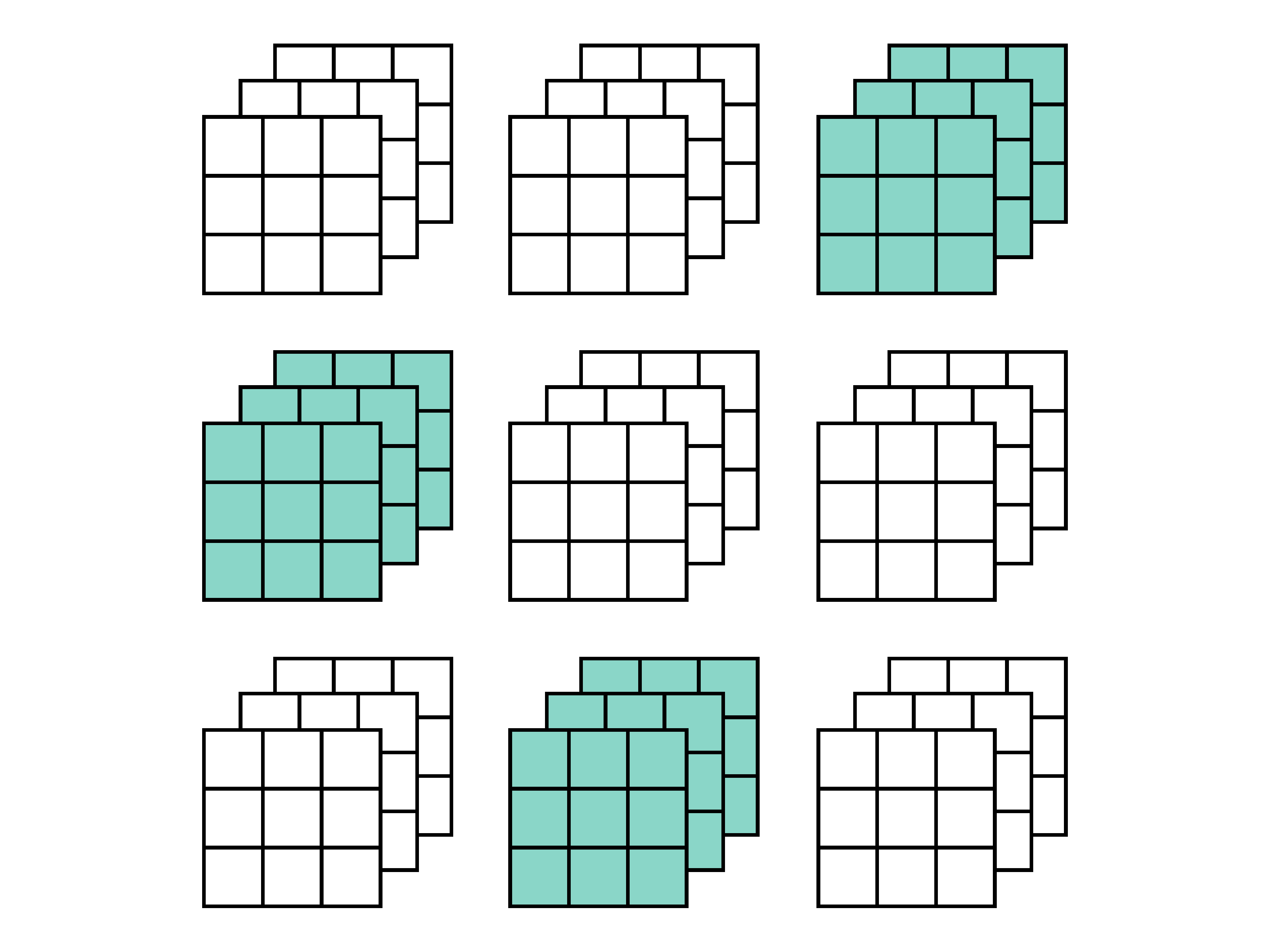}
         \caption{\texttt{\textquotesingle filter\textquotesingle}}
         \label{fig:five over x}
     \end{subfigure}
        \caption{Common granularities, available in \texttt{FasterAI}. \textcolor{myteal}{Colored} weights are selected and removed according to the chosen granularity.}
        \label{fig:granularities}        
\end{figure}

\begin{figure}[!htbp]
     \centering
     \begin{subfigure}[t]{0.24\textwidth}
         \centering
         \includegraphics[width=\textwidth]{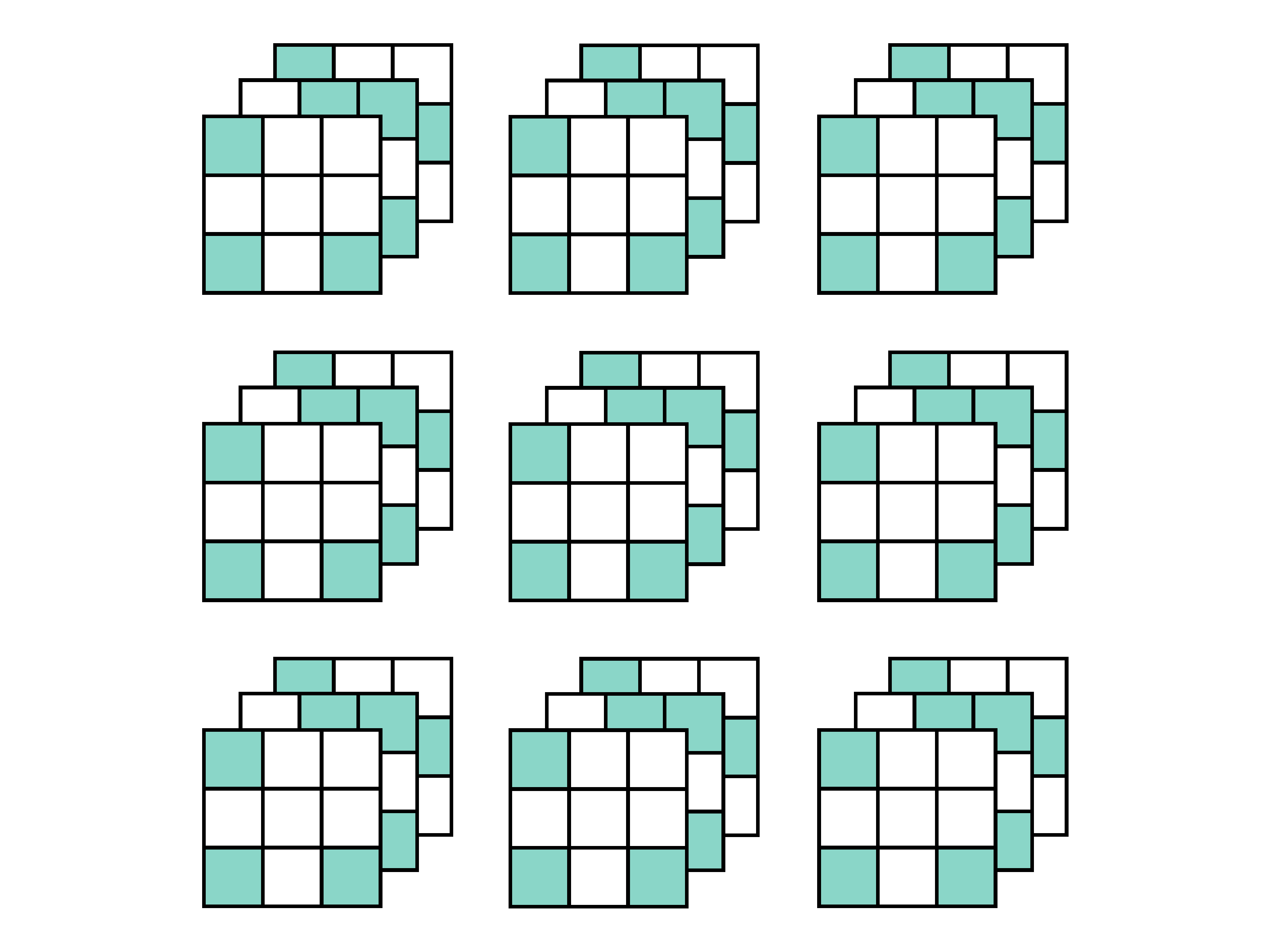}
         \caption{\texttt{\textquotesingle shared\_weight\textquotesingle}}
         \label{fig:shared_weight}
     \end{subfigure}
     \hfill
     \begin{subfigure}[t]{0.24\textwidth}
         \centering
         \includegraphics[width=\textwidth]{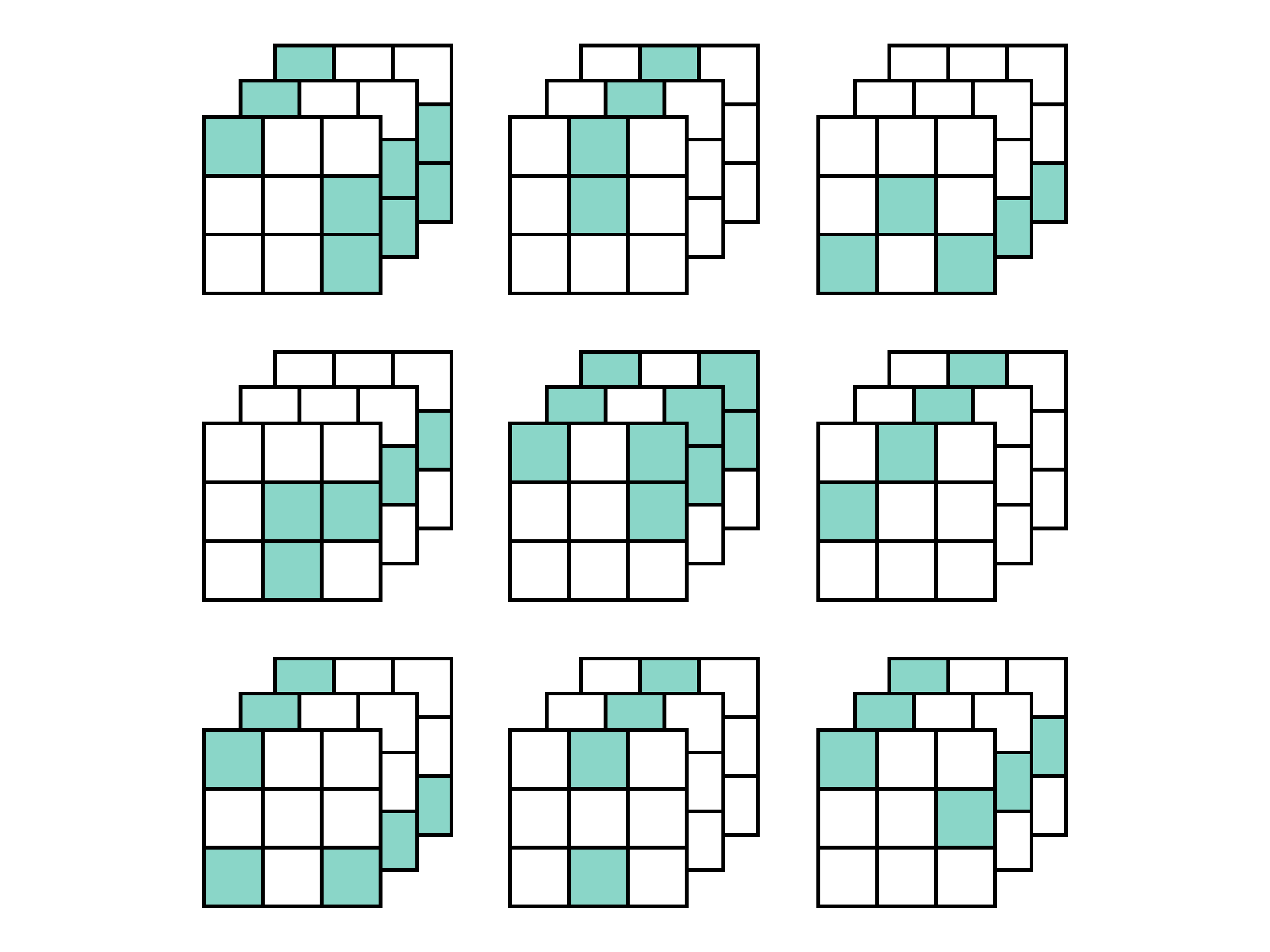}
         \caption{\texttt{\textquotesingle channel\textquotesingle}}
         \label{fig:three sin x}
     \end{subfigure}
     \hfill
     \begin{subfigure}[t]{0.24\textwidth}
         \centering
         \includegraphics[width=\textwidth]{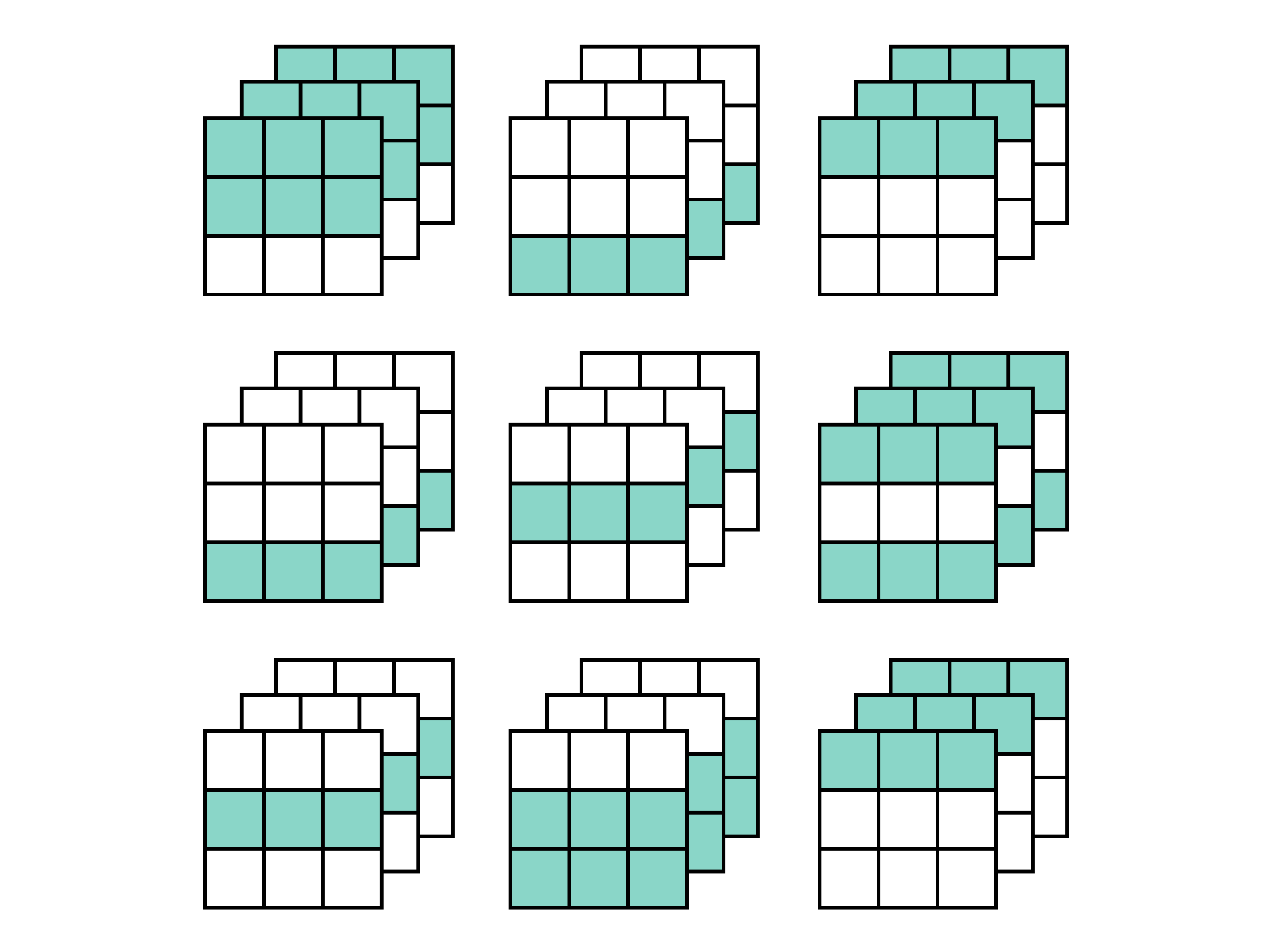}
         \caption{\texttt{\textquotesingle horizontal\_slice\textquotesingle}}
         \label{fig:five over x}
     \end{subfigure}
    \hfill
     \begin{subfigure}[t]{0.24\textwidth}
         \centering
         \includegraphics[width=\textwidth]{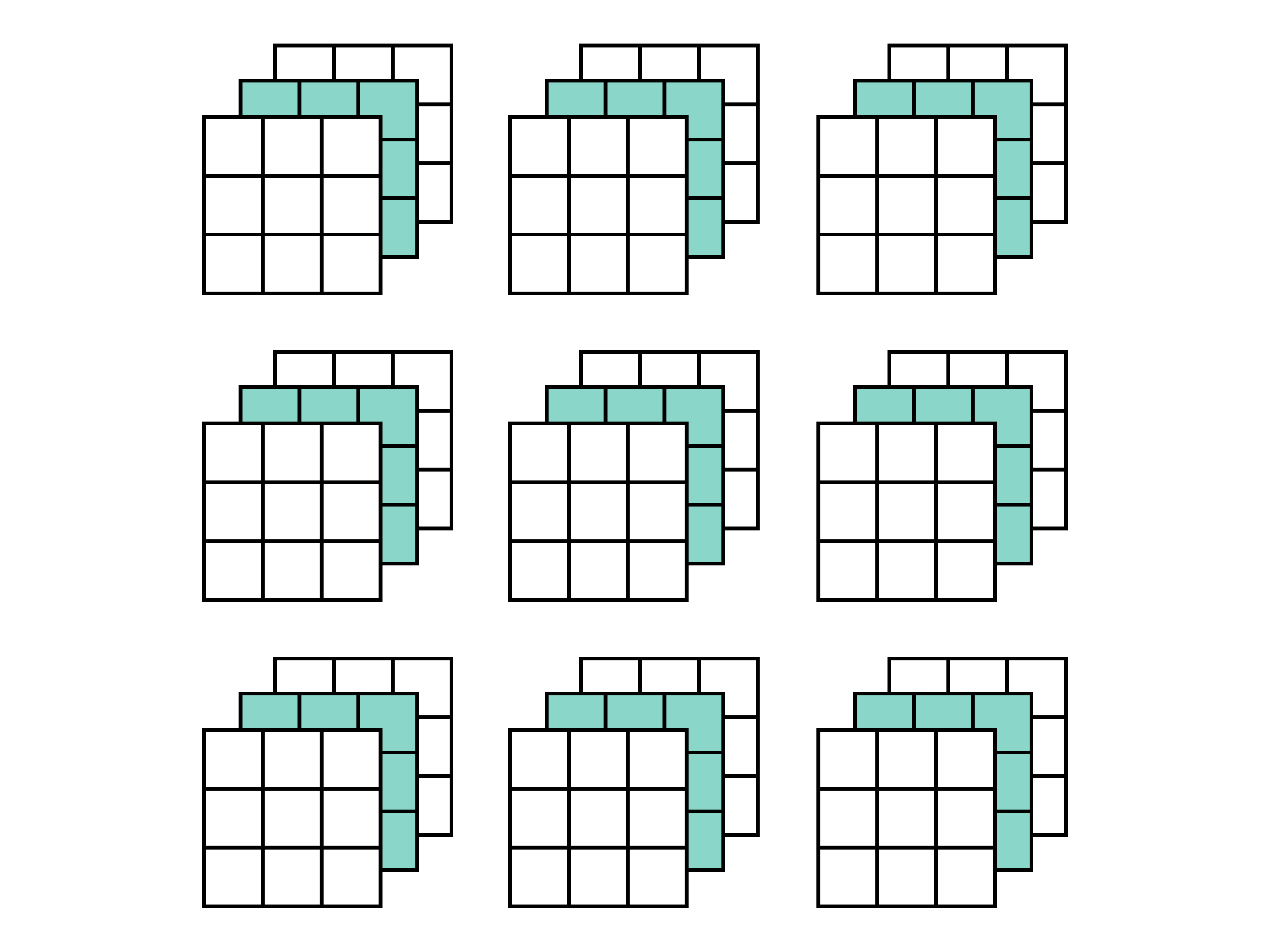}
         \caption{\texttt{\textquotesingle shared\_kernel\textquotesingle}}
         \label{fig:five over x}
     \end{subfigure}
        \caption{Uncommon granularities, available in \texttt{FasterAI}. \textcolor{myteal}{Colored} weights are selected and removed according to the chosen granularity.}
        \label{fig:granularities_unc}        
\end{figure}

In \texttt{FasterAI}, the granularity designates the structure of the blocks of weights that we want to remove. \texttt{FasterAI} handles most common sparsifying granularities, e.g. weight, kernel, filter (Figure \ref{fig:granularities}), but also allows the use of more seldom ones, e.g. horizontal slices, shared-kernels (Figure \ref{fig:granularities_unc}). In total, there are 16 granularities available for ConvNets by default in \texttt{FasterAI}, corresponding to all slicing combinations of the 4D weight tensor. Using either of those granularities comes down to passing the corresponding parameter for the \texttt{granularity} argument.

By following \texttt{PyTorch} conventions \cite{pytorch}, the weights of a Convolutional layer are given by a 4D tensor of dimension \texttt{[I, O, Kx, Ky]}, with \texttt{I, O} being respectively the input and output dimensions, and \texttt{Kx, Ky}, dimensions of the convolutional kernel. The selection of granularities displayed in Figure \ref{fig:granularities} is given by:

\begin{python}
Weight (0D) = Weights[i, o, kx, ky]
Row    (1D) = Weights[i, o, kx, :] 
Kernel (2D) = Weights[i, o, :, :] 
Filter (3D) = Weights[i, :, :, :] 
\end{python}

As previously stated, \texttt{FasterAI} allows a wide variety of granularities, along which networks parameters will be sparsified. In Figure \ref{fig:granularities_unc}, we have represented less common granularities that are available by default. For example, \texttt{shared\_weight} in Figure \ref{fig:shared_weight} defines a granularity where weights seem to be selected individually in each filter, but the same selection pattern is applied to each filter.

\begin{python}
Shared-Weight    (1D) = Weights[:, o, kx, ky]
Channel          (1D) = Weights[i, :, kx, ky] 
Horizontal-Slice (2D) = Weights[i, :, kx,:] 
Shared-Kernel    (3D) = Weights[:, o, :, :] 
\end{python}

\begin{figure}[!htbp]
     \centering
     \begin{subfigure}[t]{0.3\textwidth}
         \centering
         \includegraphics[width=0.7\textwidth]{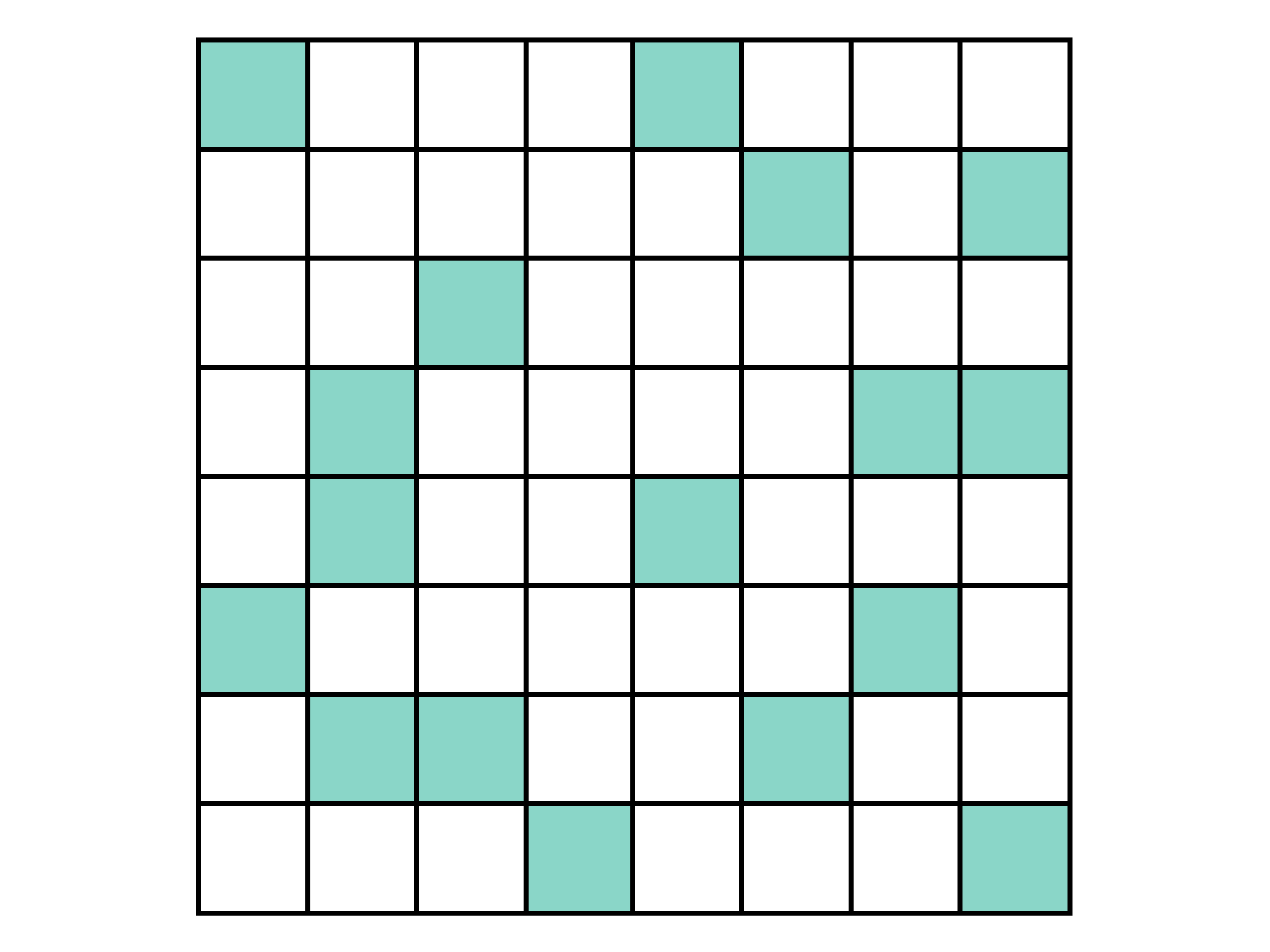}
         \caption{\texttt{\textquotesingle weight\textquotesingle}}
         \label{fig:y equals x}
     \end{subfigure}
     \hfill
     \begin{subfigure}[t]{0.3\textwidth}
         \centering
         \includegraphics[width=0.7\textwidth]{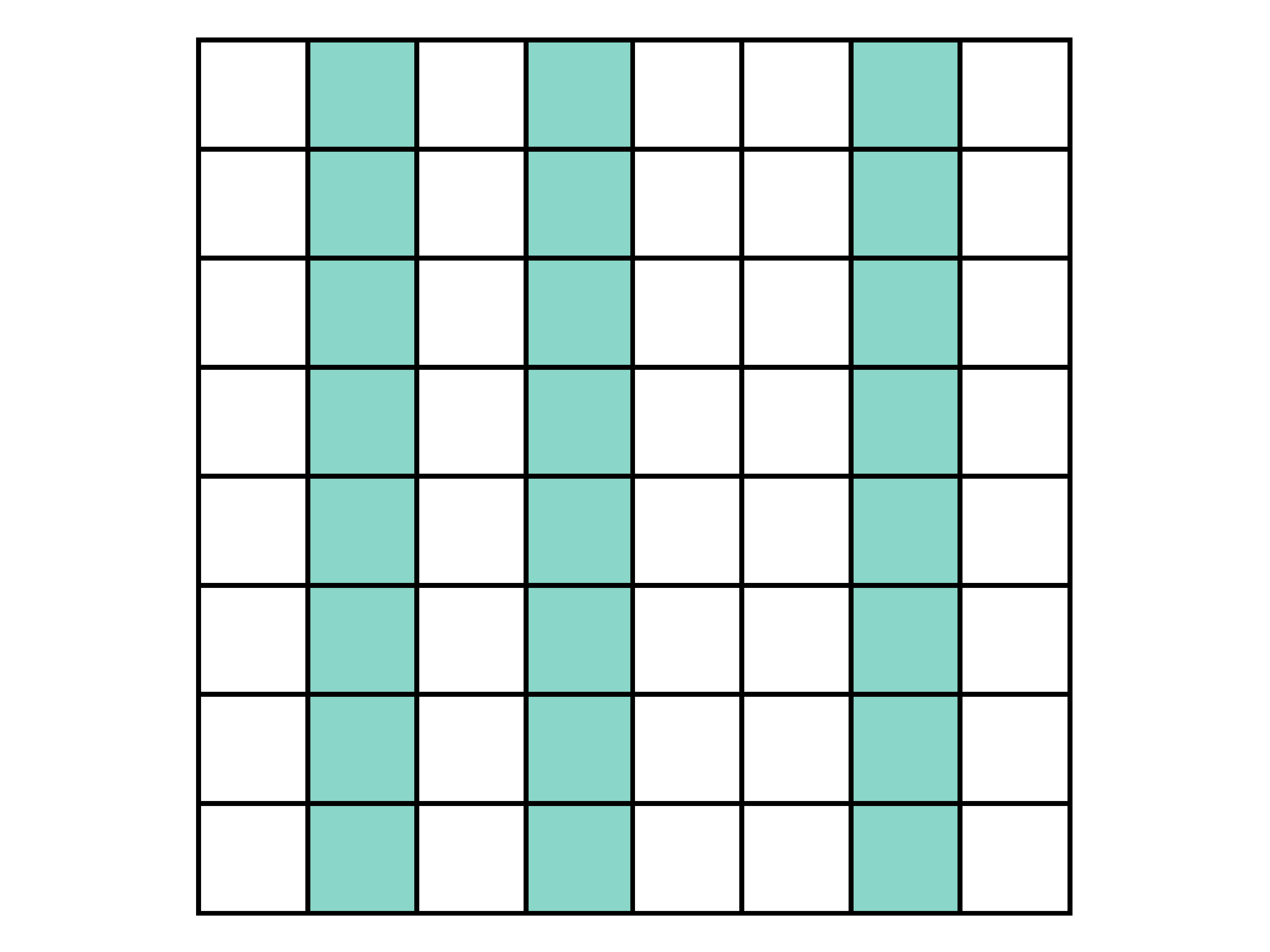}
         \caption{\texttt{\textquotesingle column\textquotesingle}}
         \label{fig:three sin x}
     \end{subfigure}
     \hfill
     \begin{subfigure}[t]{0.3\textwidth}
         \centering
         \includegraphics[width=0.7\textwidth]{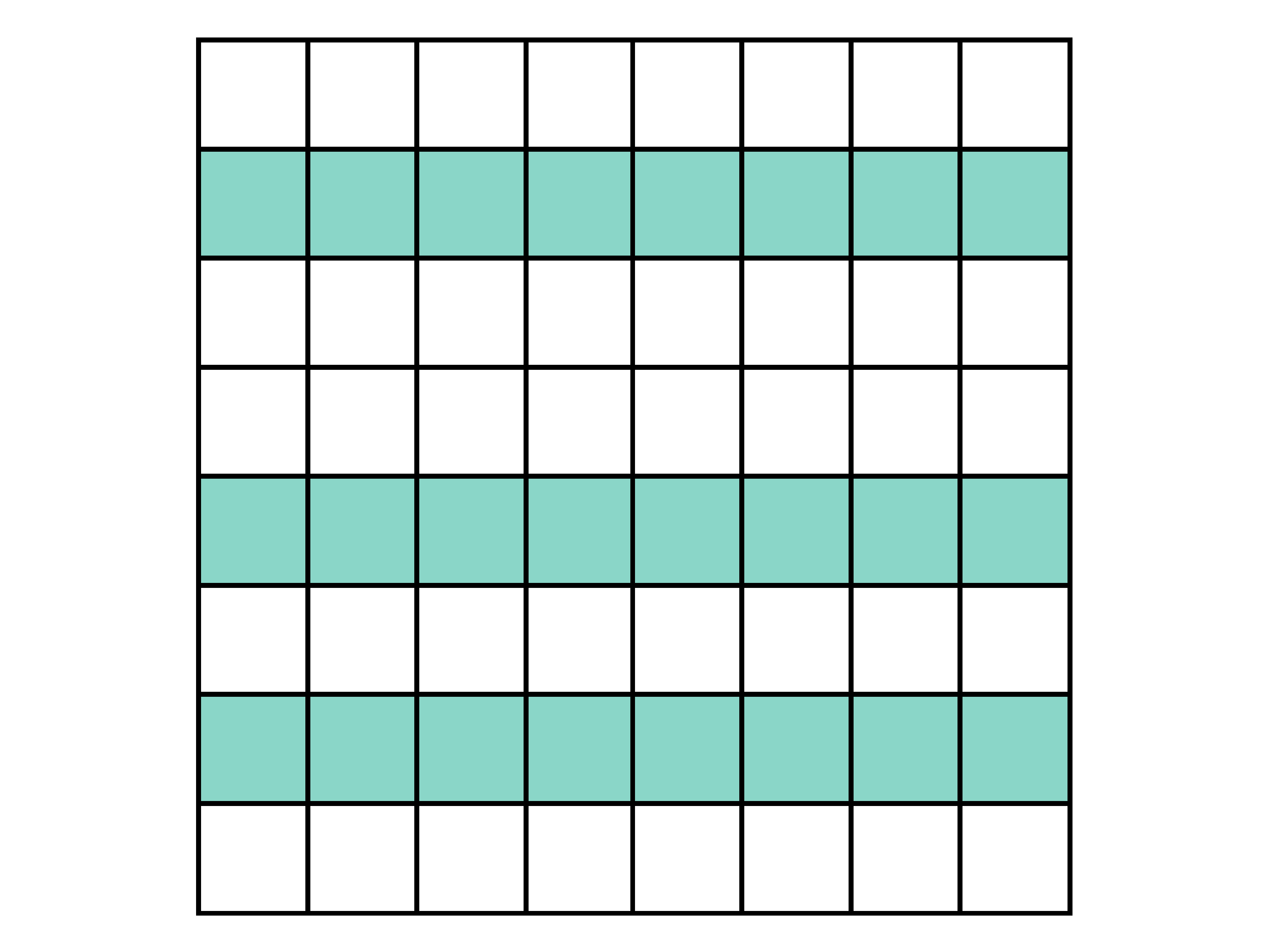}
         \caption{\texttt{\textquotesingle row\textquotesingle}}
         \label{fig:five over x}
     \end{subfigure}
    \hfill
        \caption{Granularities available for Fully-Connected Layers. \textcolor{myteal}{Colored} weights are selected and removed according to the chosen the granularity.}
        \label{fig:granularities_linear}        
\end{figure}

Granularities for other types of layers such as Fully-Connected or Attention layers are also available, but the options are more limited due to the weight matrix being of a smaller rank. Available granularities for Fully-Connected layers are represented in Figure \ref{fig:granularities_linear}.

The corresponding Pseudo-Codes to obtain such granularities from a 2D weight matrix \texttt{[I,O]} are given below:

\begin{python}
Weight (0D) = Weights[i, o]
Column (1D) = Weights[:, o]
Row    (1D) = Weights[i, :]
\end{python}

\subsection{\texttt{context}: where to sparsify?}

In \texttt{FasterAI}, the context refers to the locality of the selection of the weights. In literature, the two most important options are: (1) local pruning, \textit{i.e.} the selection of the weights is performed in each layer separately, producing equally sparse layers in the network, and (2) global pruning, \textit{i.e.} the selection of the weights is performed by comparing those of the whole network, producing a network with different sparsity levels for each layer. Although \texttt{FasterAI} can handle both methods by passing either \texttt{\textquotesingle local\textquotesingle} or \texttt{\textquotesingle global\textquotesingle} to the callback, Pseudo-Code for how to perform both techniques is given below:

\begin{python}
# (1) Local Pruning
for layer in layers:
	mask = compute_mask(layer.weight, sparsity)
	pruned_model = prune_layer(layer, mask)
	
# (2) Global Pruning
global_weights = concat[(layer.weight) for layer in layers]
mask = compute_mask(global_weights, sparsity)
pruned_model = prune_model(mask)
\end{python}

	

Local and global pruning have very different implications on the final sparsity of the network, with local pruning leading to equally sparse layers all along the network and global pruning leading to layers with sometimes very large differences in sparsities. In case user wants to specify a particular sparsity level for certain layers, \texttt{FasterAI} accepts a list of sparsities, that will be applied to each corresponding layer. 

\subsection{\texttt{criteria}: what to sparsify?}

\begin{figure}[!htbp]
     \centering
     \begin{subfigure}[t]{0.24\textwidth}
         \centering
         \includegraphics[width=0.9\textwidth]{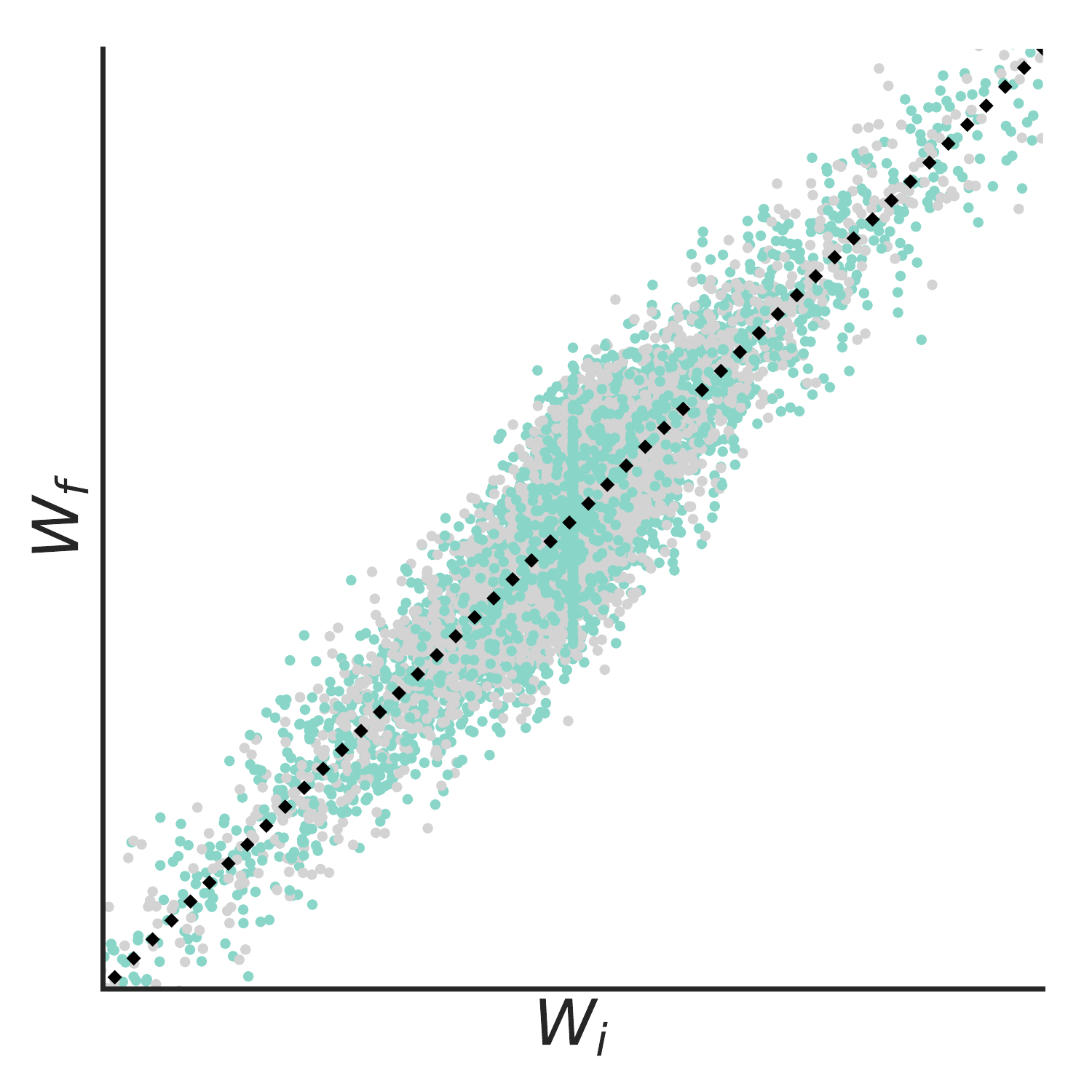}
         \caption{\texttt{random}}
         \label{fig:y equals x}
     \end{subfigure}
     \hfill
     \begin{subfigure}[t]{0.24\textwidth}
         \centering
         \includegraphics[width=0.9\textwidth]{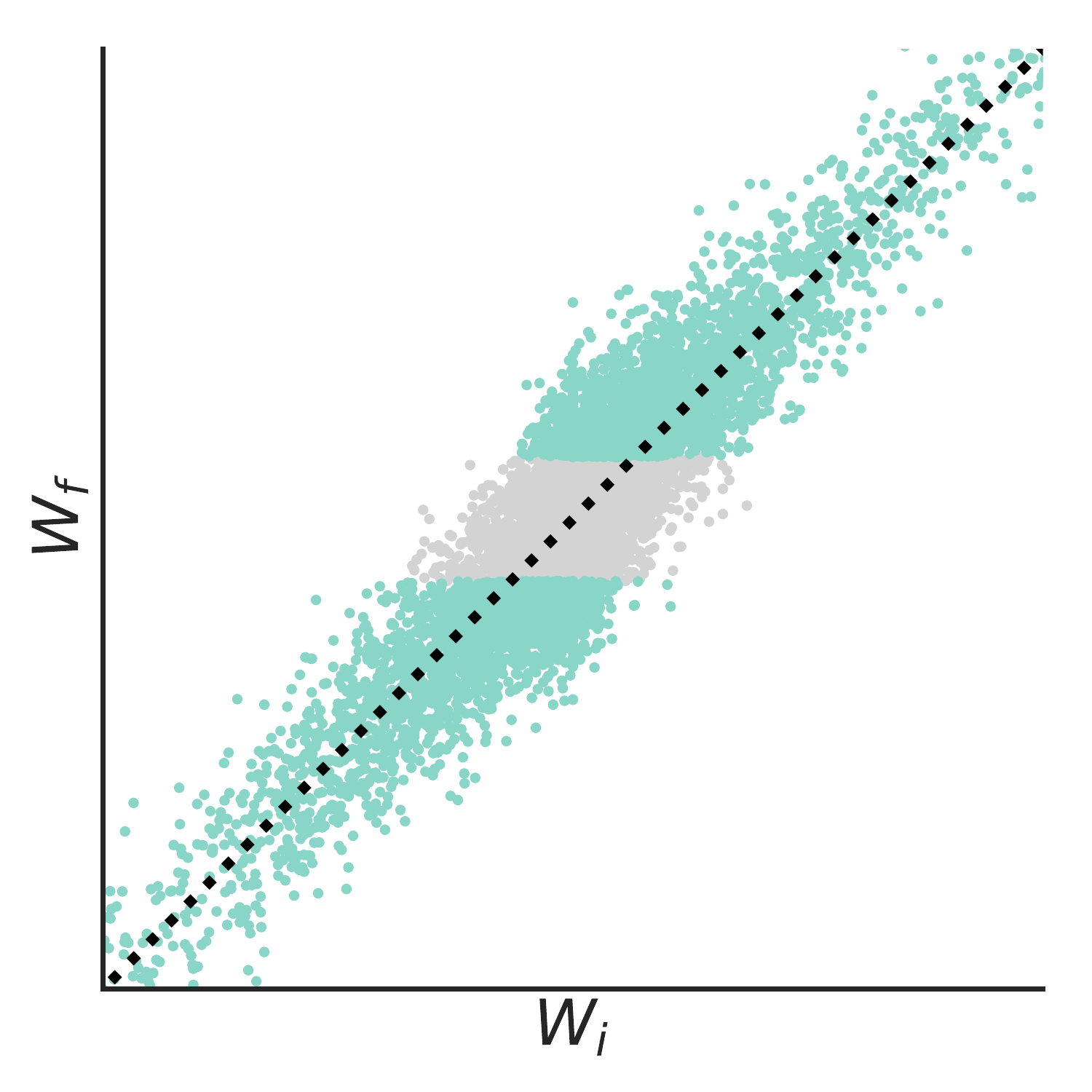}
         \caption{\texttt{large\_final}}
         \label{fig:three sin x}
     \end{subfigure}
     \hfill
     \begin{subfigure}[t]{0.25\textwidth}
         \centering
         \includegraphics[width=0.9\textwidth]{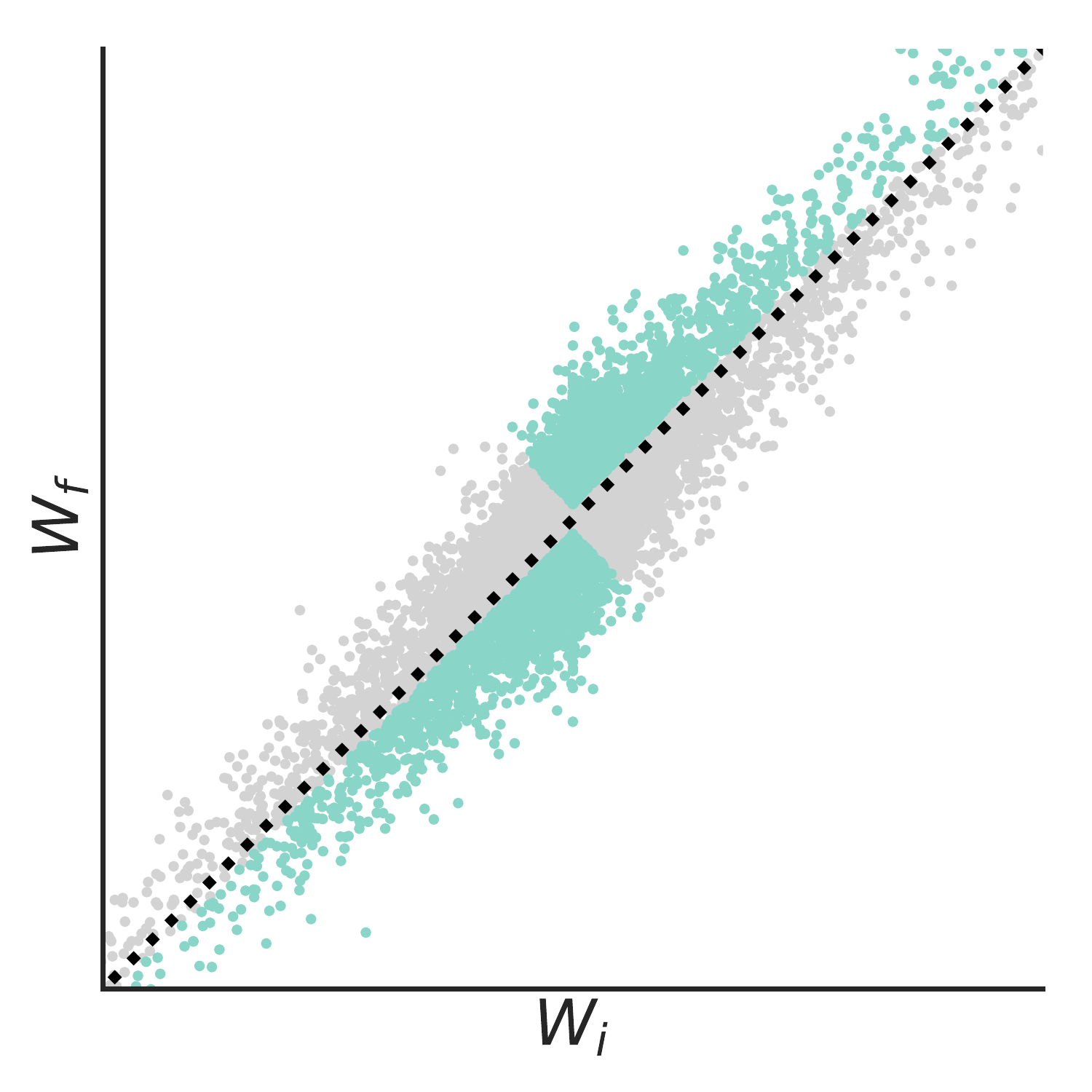}
         \caption{\texttt{magnitude\_increase}}
         \label{fig:five over x}
     \end{subfigure}
    \hfill
     \begin{subfigure}[t]{0.24\textwidth}
         \centering
         \includegraphics[width=0.9\textwidth]{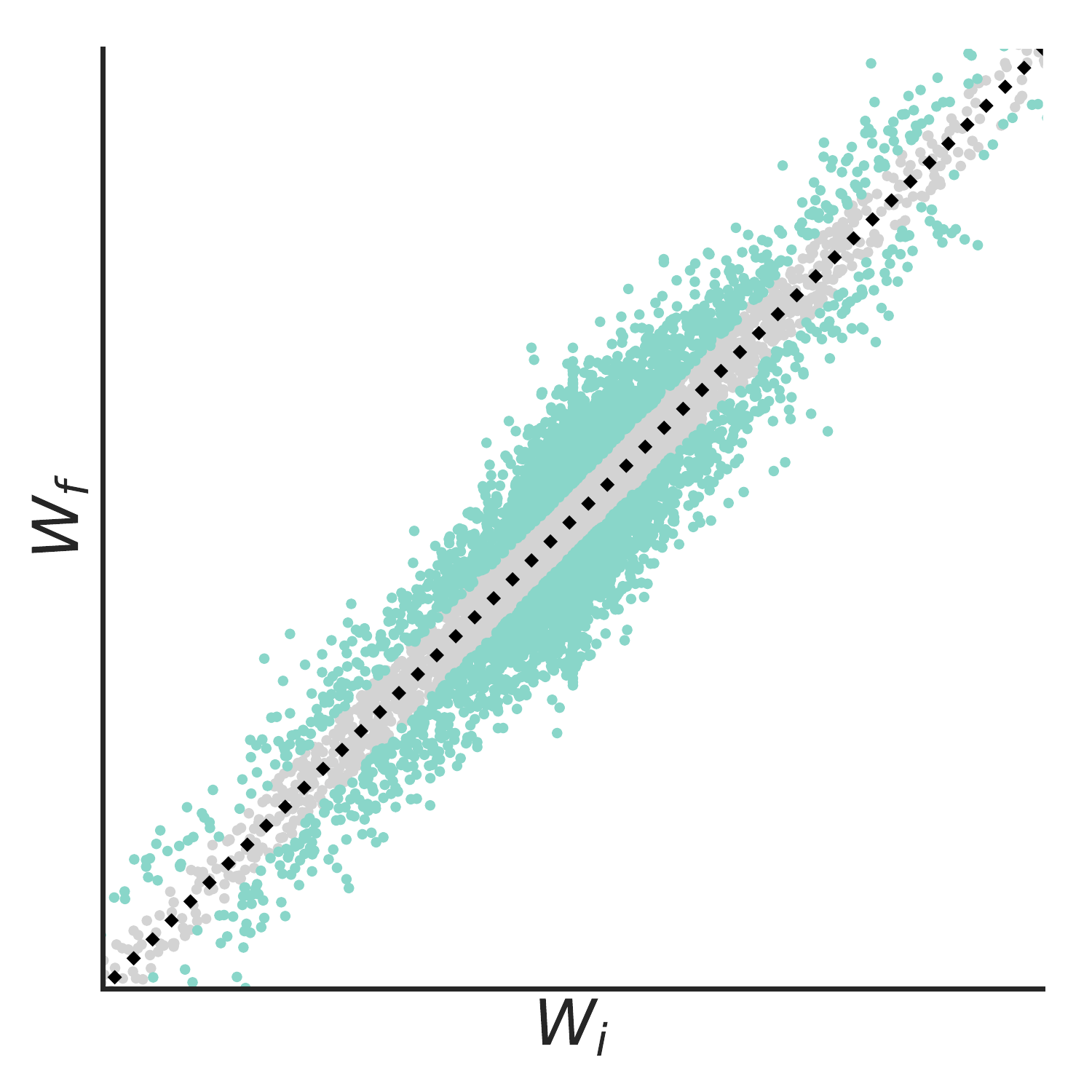}
         \caption{\texttt{movement}}
         \label{fig:five over x}
     \end{subfigure}
        \caption{Common pruning selection criteria. A schematic weight distribution is represented, where \textcolor{myteal}{colored} weights are considered as important by the criteria, while greyed out ones are removed.}
        \label{fig:criteria}        
\end{figure}

In \texttt{FasterAI}, we designate by criteria the score according to which the importance of parameters will be assessed. \texttt{FasterAI} currently lists 13 different criteria. The most common ones are displayed in Figure \ref{fig:criteria}, following the representation and naming conventions of Zhou et al. \cite{supermasks}, where the x-axis represents values of weights at the previous pruning iteration ($W_i$) and y-axis represents weights at the current iteration ($W_f$).

The Pseudo-Code to select weights for the represented criteria is given below:

\begin{python}
random             = torch.randn_like(wf)
large_final        = torch.abs(wf)
magnitude_increase = torch.sub(torch.abs(wf),torch.abs(wi))
movement           = torch.abs(torch.sub(wf,wi))
\end{python}

In practice, the criteria is applied to each weight, before aggregating them according to the \texttt{granularity}. The pruning mask is then computed by taking the weights having the largest score, according to the desired \texttt{context} and sparsity level. \texttt{FasterAI} keeps track of previous values of the weights, allowing for pruning criteria such as Magnitude Increase or Movemement, which rely on comparing weight values at the current training iteration ($W_f$) to the values they had at the previous iteration ($W_i$). This enables the creation of pruning criteria that take training dynamics into account.


\subsection{\texttt{schedule}: when to sparsify?}

\texttt{FasterAI} allows the user to create or customize its own pruning schedules by adjusting 3 parameters in the \texttt{SparsifyCallback}: 

\begin{itemize}
\item \texttt{start\_epoch} (default to \texttt{0}): the epoch at which we want the pruning to start, \textit{i.e.} for how many epochs we want to pretrain our model.
\item \texttt{end\_epoch} (default to \texttt{total\_epoch}): the epoch at which we want the pruning to end, \textit{i.e.} for how many epochs we want to fine-tune our model after pruning.
\item \texttt{schedule\_function}: the function describing the evolution of the sparsity during the training. There are 4 currently available: One-Shot, Iterative, Gradual \cite{agp}, and One-Cycle \cite{onecycle}.
\end{itemize}

Here are for example, the 4 available pruning schedules:

\begin{figure}[!htbp]
     \centering
     \begin{subfigure}[t]{0.24\textwidth}
         \centering
         \includegraphics[width=\textwidth]{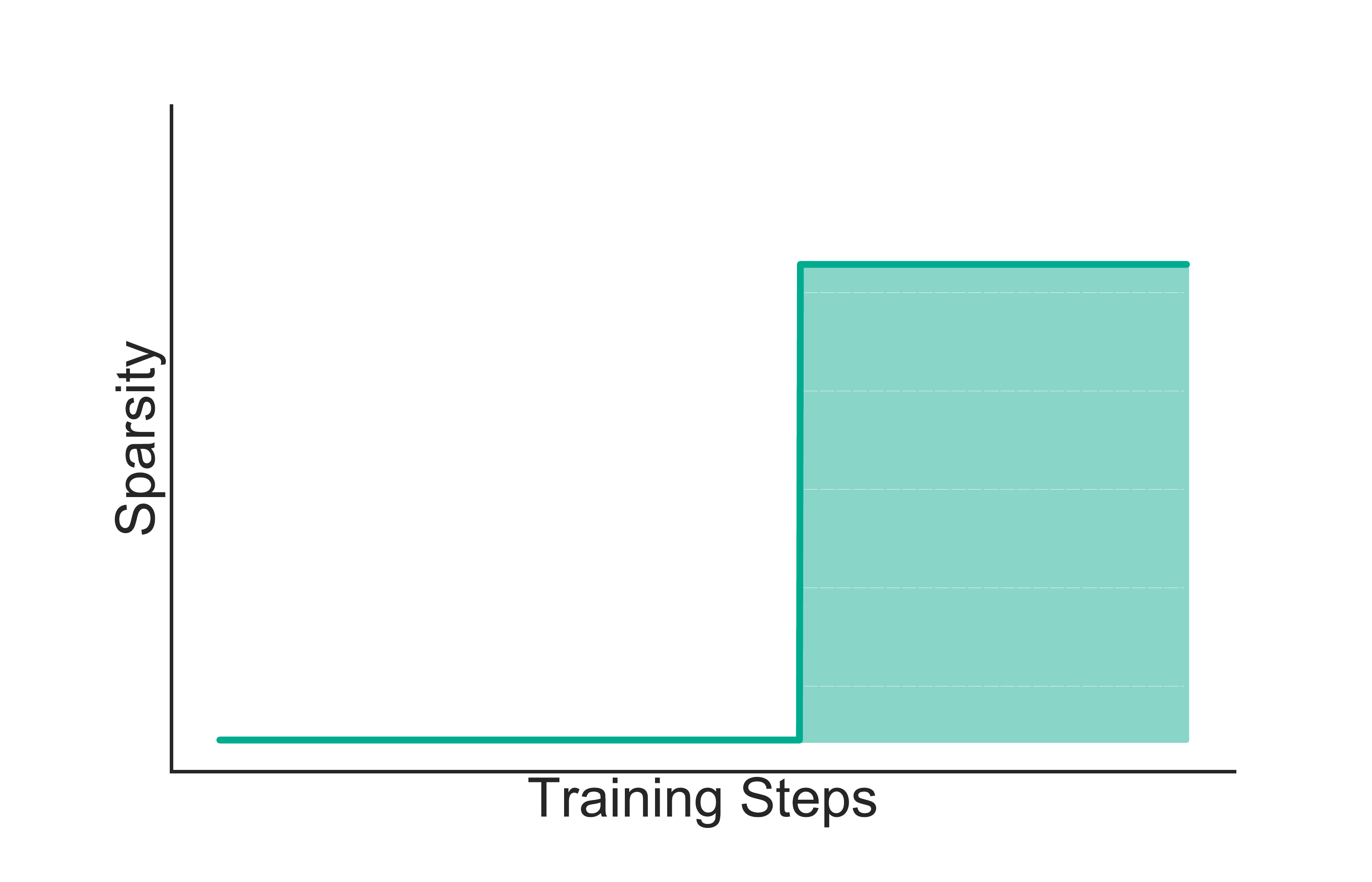}
         \caption{\texttt{one\_shot}}
         \label{fig:one_shot}
     \end{subfigure}
     \hfill
     \begin{subfigure}[t]{0.24\textwidth}
         \centering
         \includegraphics[width=\textwidth]{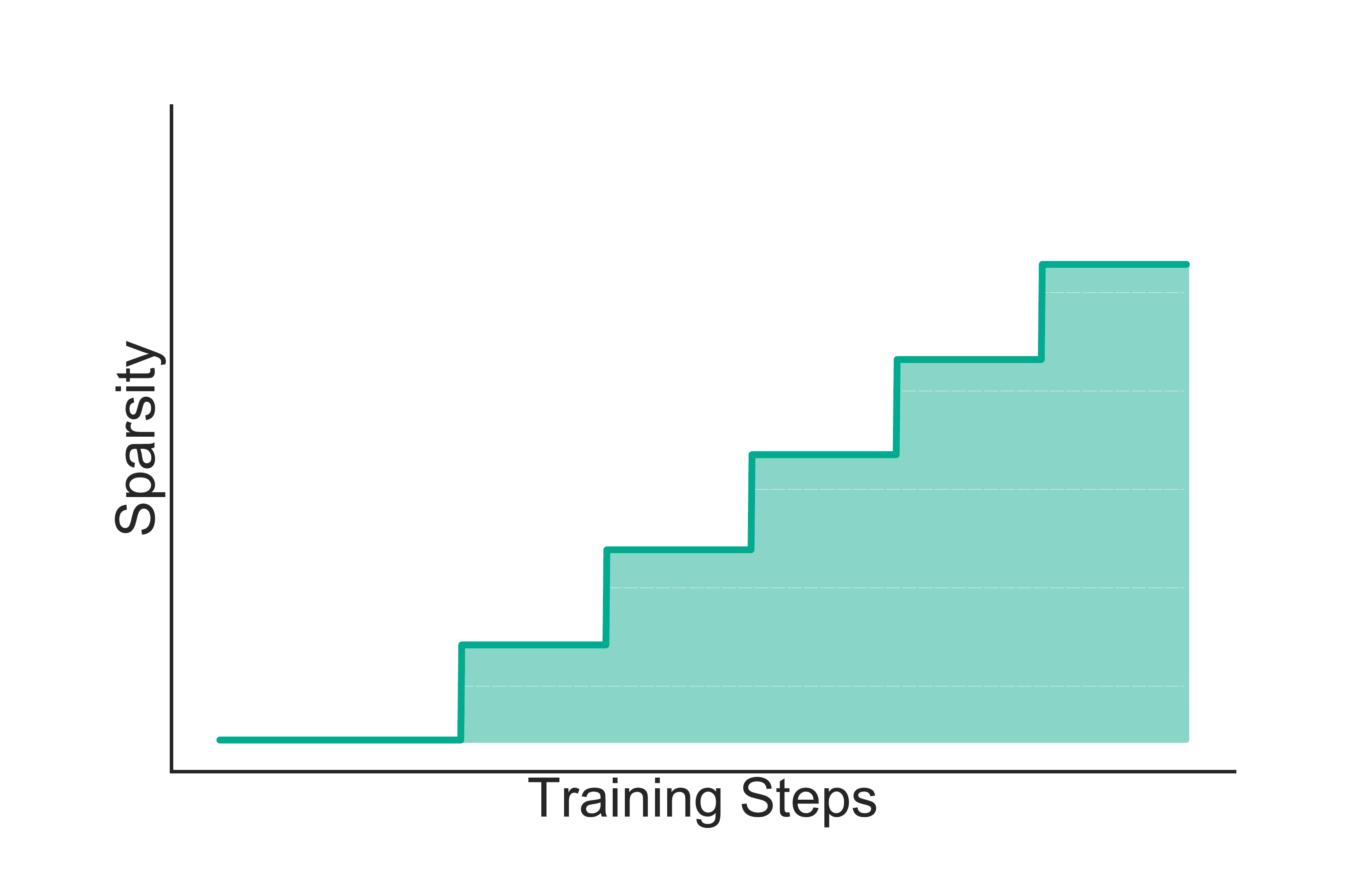}
         \caption{\texttt{iterative}}
         \label{fig:three sin x}
     \end{subfigure}
     \hfill
     \begin{subfigure}[t]{0.24\textwidth}
         \centering
         \includegraphics[width=\textwidth]{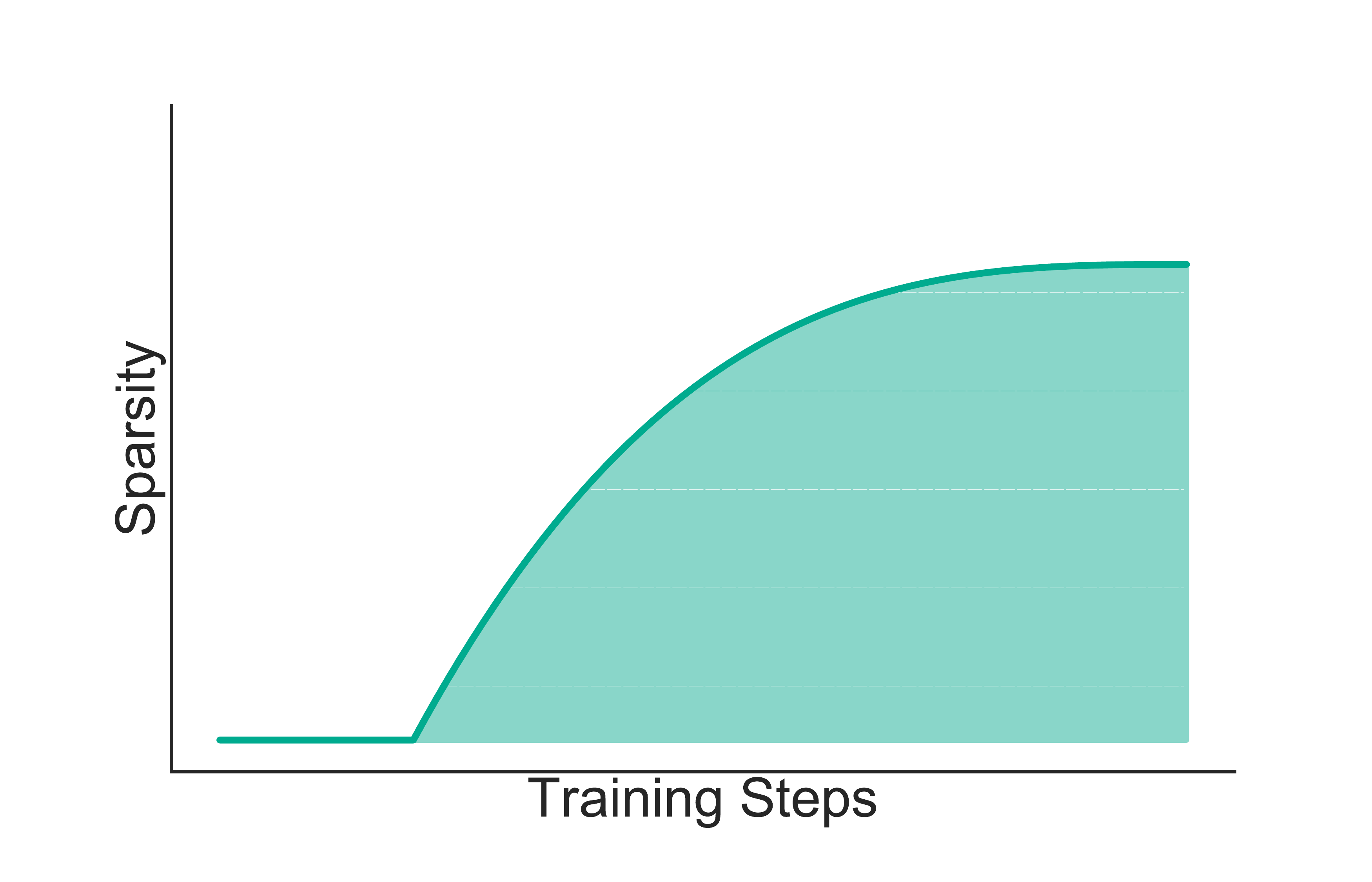}
         \caption{\texttt{gradual}}
         \label{fig:five over x}
     \end{subfigure}
    \hfill
     \begin{subfigure}[t]{0.24\textwidth}
         \centering
         \includegraphics[width=\textwidth]{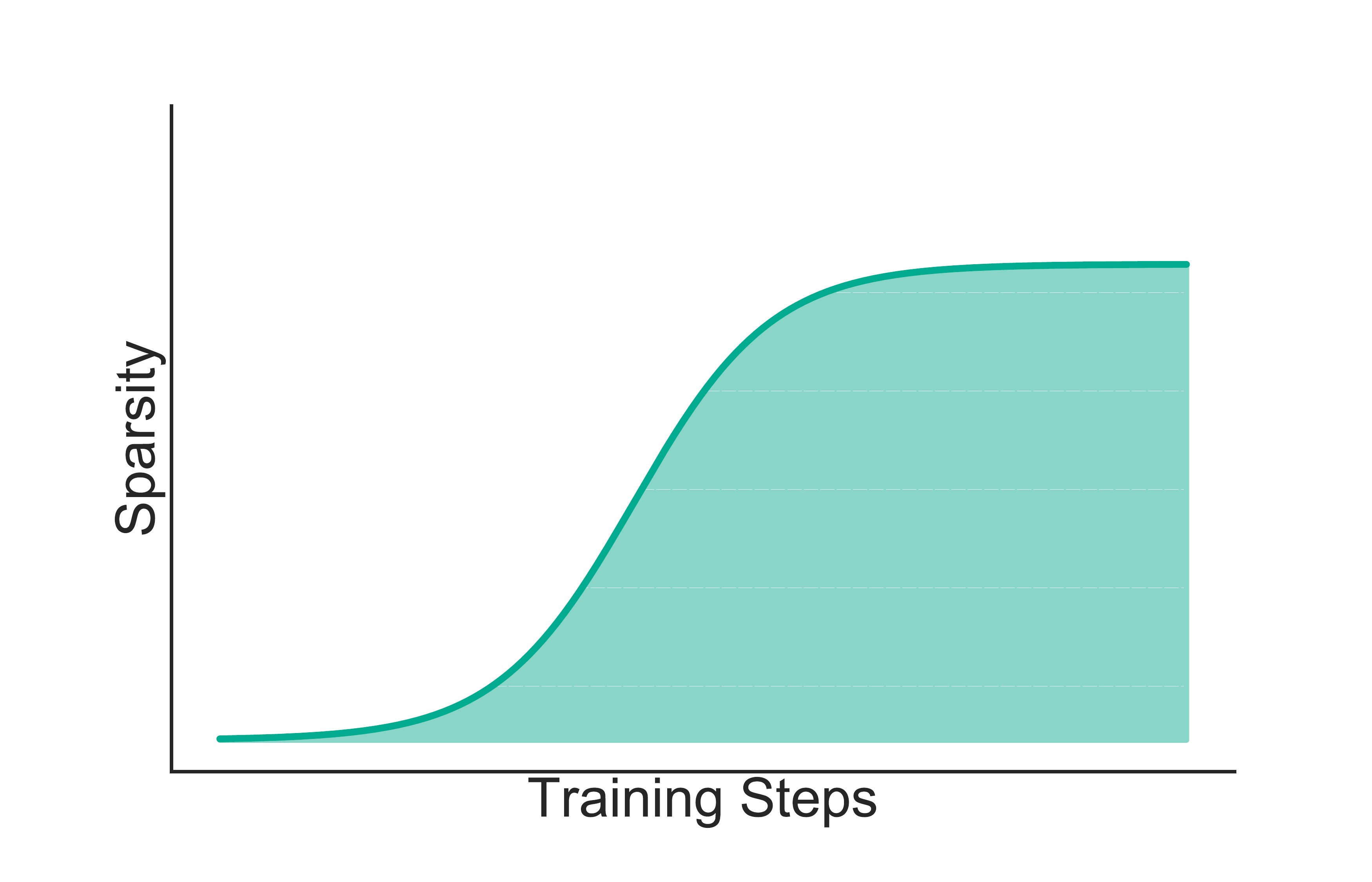}
         \caption{\texttt{one\_cycle}}
         \label{fig:five over x}
     \end{subfigure}
        \caption{Evolution of sparsity for the available pruning schedules.}
        \label{fig:Schedules}        
\end{figure}

All of them are easily obtained in \texttt{FasterAI}. Here are the pseudo-codes for Figure \ref{fig:Schedules}.

\begin{python}
def one_shot(sparsity, t_step): return sparsity

def iterative(sparsity, t_step, n_steps=5):
    return  (sparsity/n_steps)*(np.ceil((t_step)*n_steps))

def gradual(sparsity, t_step): return sparsity * (1 - t_step)**3

def one_cycle(sparsity, t_step, $\alpha$=14, $\beta$=6):
    return (1+np.exp(-$\alpha$+$\beta$)) / (1 + (np.exp(-$\alpha$*t_step+$\beta$)))*sparsity

\end{python}

The \texttt{start\_epoch} and \texttt{end\_epoch} can further help the user to alter the pruning schedule as desired. For example, in Figure \ref{fig:one_shot}, the One-Shot pruning schedule could be used with a value of \texttt{start\_epoch=0}, becoming what is more well-known as Pruning at Initialization \cite{pai}, achieving the target amount of sparsity right from the start of training.


By modifying those parameters, users can create their own pruning schedule or easily implement other existing ones, such as the dense-sparse-dense (DSD) schedule \cite{DSD} for example, which increases the sparsity for the first half of training, then gradually decay it until the network is $0\%$ sparse again: 

\begin{python}
def dsd(sparsity, t_step):
    if t_step<0.5: return (1 + math.cos(math.pi*(1-t_step*2))) * sparsity / 2
    else: return (1 - math.cos(math.pi*(1-t_step*2))) * sparsity / 2
\end{python}

\begin{figure}[!htbp]
     \centering
     \begin{subfigure}[t]{0.3\textwidth}
         \centering
         \includegraphics[width=\textwidth]{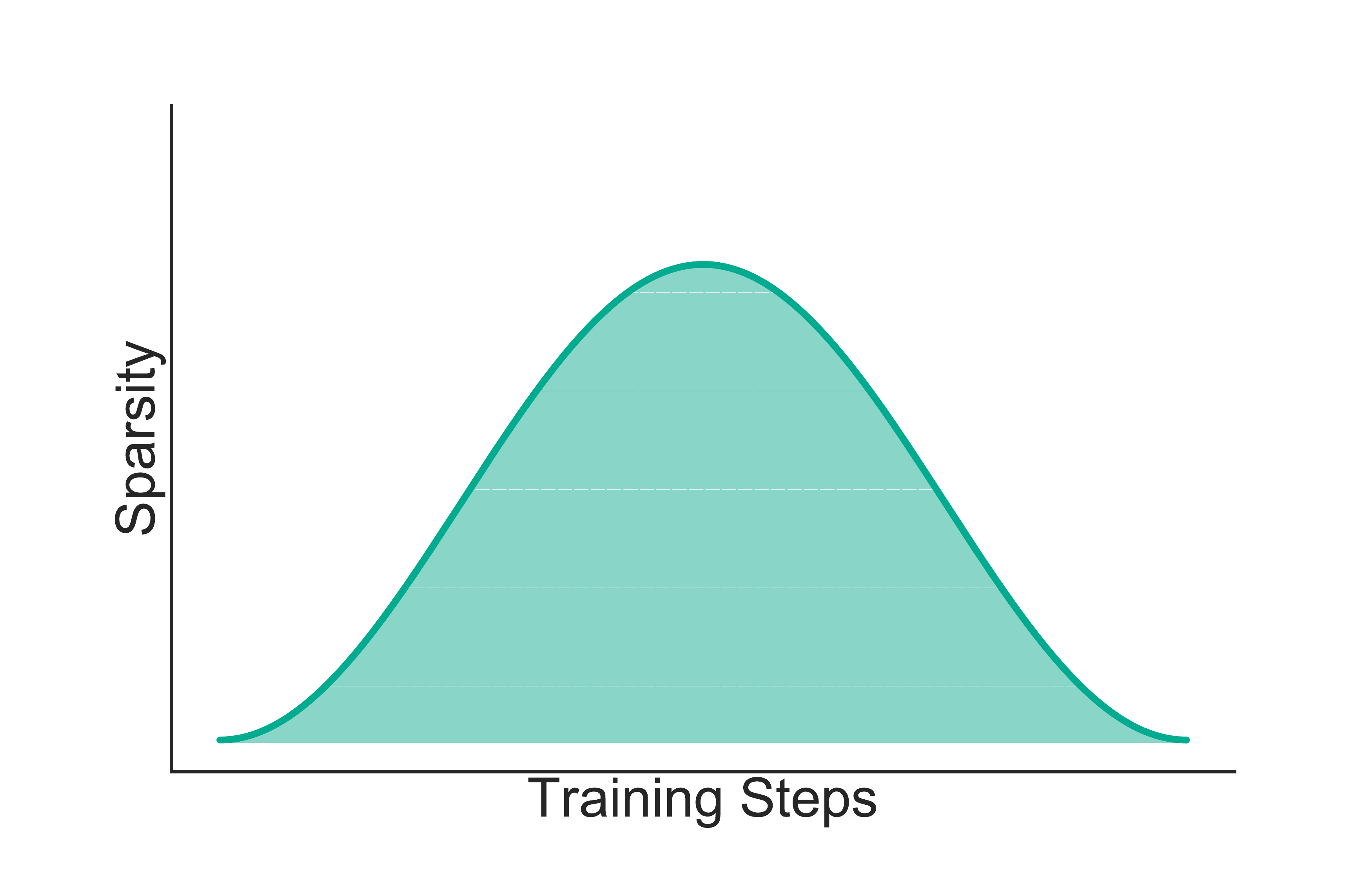}
         \caption{\texttt{dsd}, \texttt{start\_epoch=0}}
         \label{fig:y equals x}
     \end{subfigure}
     \hfill
     \begin{subfigure}[t]{0.3\textwidth}
         \centering
         \includegraphics[width=\textwidth]{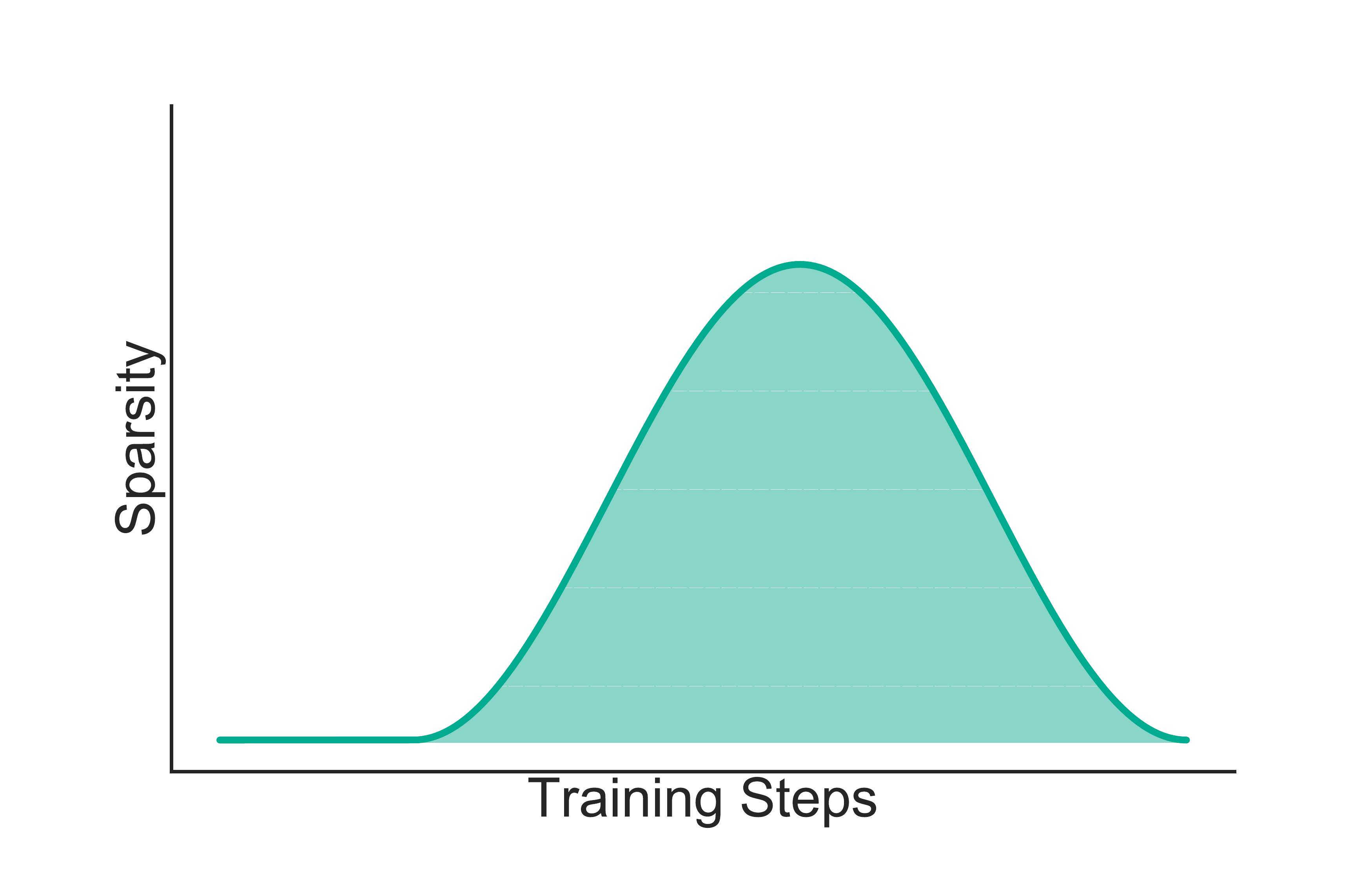}
         \caption{\texttt{dsd}, \texttt{start\_epoch>0}}
         \label{fig:three sin x}
     \end{subfigure}
     \hfill
     \begin{subfigure}[t]{0.32\textwidth}
         \centering
         \includegraphics[width=\textwidth]{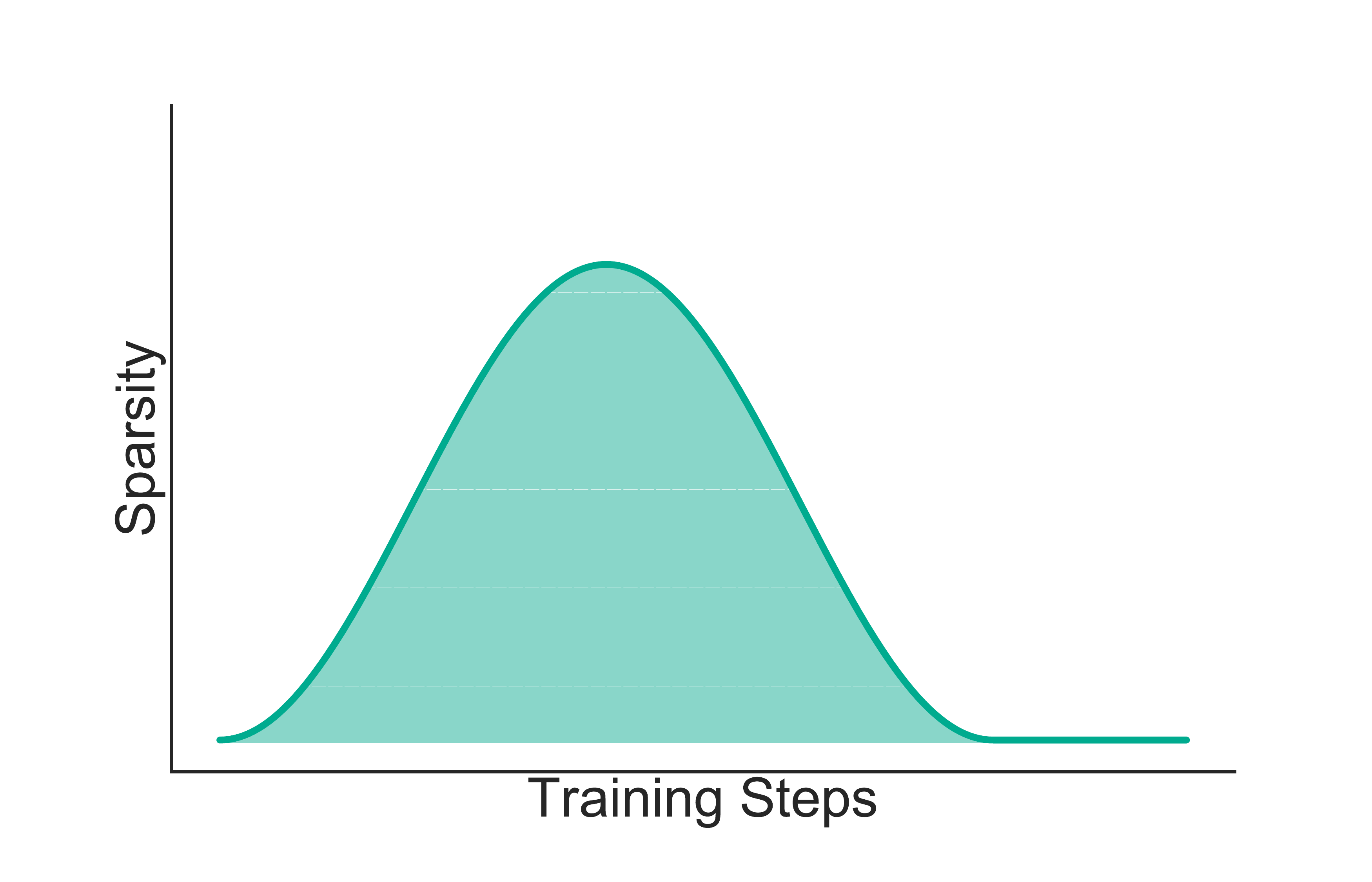}
         \caption{\texttt{dsd}, \texttt{end\_epoch<total\_epoch}}
         \label{fig:five over x}
     \end{subfigure}
        \caption{Variation of the \texttt{dsd} schedule. The use of \texttt{start\_epoch} and \texttt{end\_epoch} help to further customize a given pruning schedule.}
        \label{fig:DSD}        
\end{figure}

By then modifying the values of \texttt{start\_epoch} and \texttt{end\_epoch} in the \texttt{SparifyCallback}, we can further customize our pruning schedule, as displayed in Figure \ref{fig:DSD}.

\subsection{Lottery Ticket Hypothesis}

\texttt{FasterAI} also handles Lottery Tickets Hypothesis (LTH) experiments by default. Those experiments generally consist of 5 steps, represented in Figure \ref{fig:LTH}: 

\begin{figure}[!htbp]
 \centering
  \centerline{\includegraphics[scale=.22]{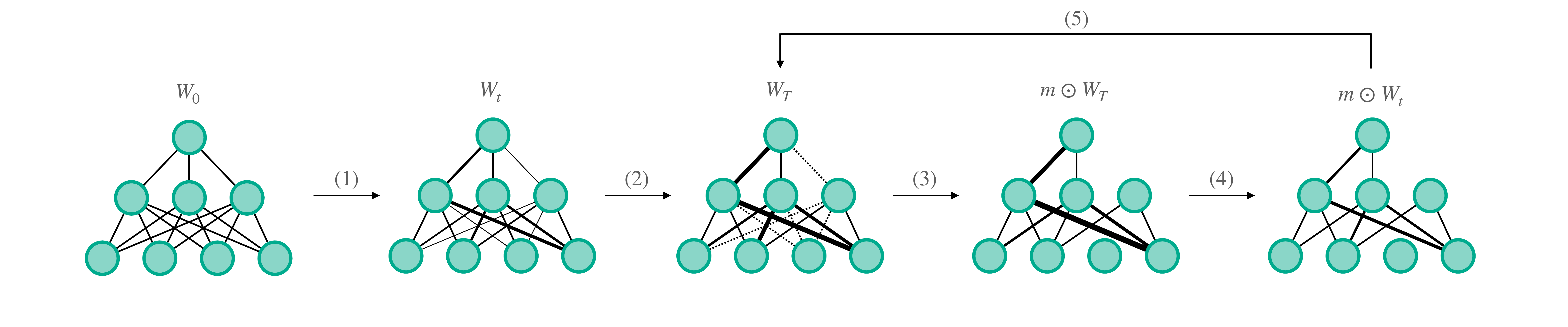}}
  \caption{The Lottery Ticket experiment.}
  \label{fig:LTH}
\end{figure}

\begin{enumerate}
    \item Train a freshly initialized network ($W_0$) for $t$ iterations and save its set of weight ($W_t$).
    \item Continue the training until completion ($W_T$).
    \item Apply a pruning mask according to the desired sparsity level, granularity, context, criteria ($m \odot W_T$).
    \item Reset the weights to their previously saved values, still applying the pruning mask ($m \odot W_t$).
    \item Continue training and repeat the previous steps, each time updating the mask until desired sparsity is achieved.
\end{enumerate}

It is worth noting that when the iteration $t$ in step 1 is set to $0$, then the procedure turns into the classic Lottery Ticket Experiment \cite{lottery}, otherwise it corresponds to the Lottery Ticket with Rewinding Experiment \cite{lottery2}. To accomplish such procedure in \texttt{FasterAI}, some additional arguments can be provided to the \texttt{SparsifyCallback}.

\begin{itemize}
    \item \texttt{lth}: whether weights are reinitialized to their saved value after each pruning round.
    \item \texttt{rewind\_epoch} (default to \texttt{0}): the epoch of training where weights values are saved for further reinitialization.
    \item \texttt{reset\_end}: whether to reset the weights to their saved values after training.
    \item \texttt{save\_tickets}: saving intermediate tickets after each pruning round.
\end{itemize}

Performing the classic Lottery Ticket Experiments \cite{lottery, lottery2} following the original pruning settings would thus look like: 

\begin{python}
# Classic LTH
SparsifyCallback(sp,'weights','global', large_final, iterative, lth=True)

# LTH with Rewinding
SparsifyCallback(sp, 'weights', 'global', large_final, iterative, lth=True,\
rewind_epoch=1)
\end{python}

As the LTH-related arguments are optional arguments, it means that one may perform LTH experiments with any granularity, context, criteria and schedules, opening the way to many novel experiments about finding winning tickets.

\section{Conclusion \& Future Development}

\texttt{FasterAI} provides a lightweight framework enabling quick and diverse experimentations on sparsifying techniques. We believe that the way the library was built laid solid foundations to allow an easier implementation of novel compression techniques. We also hope that such a library, because it was built in a granular way, allowing to easily combine all different sorts of pruning techniques, will help researchers to perform new kinds of experiments. With this goal in mind, other compression capabilities are also available and further developed in \texttt{FasterAI}, such as knowledge distillation, regularization, which can also be used in combination with sparsification techniques for even more original experimentations.

\bibliographystyle{IEEEbib}
\bibliography{biblio}

\end{document}